\documentclass[journal]{IEEEtran}
\usepackage{graphicx} 
\usepackage{svg}
\begin{document}

\title{Large Language Models for Wireless Communications: From Adaptation to Autonomy}
\author{Le Liang, Hao Ye, Yucheng Sheng, Ouya Wang, Jiacheng Wang, Shi Jin, and Geoffrey Ye Li

\thanks{The work was supported by the National Key R\&D Program of China under Grant 2024YFE0200700.}
\thanks{L. Liang, Y. Sheng, J. Wang, and S. Jin (corresponding author) are with Southeast University, China; H. Ye is with the University of California, Santa Cruz, USA; O. Wang and G. Y. Li are with the Imperial College London, UK.}
}
\maketitle
\begin{abstract}
The emergence of large language models (LLMs) has revolutionized artificial intelligence, offering unprecedented capabilities in reasoning, generalization, and zero-shot learning. These strengths open new frontiers in wireless communications, where increasing complexity and dynamics demand intelligent and adaptive solutions. This article explores the role of LLMs in transforming wireless systems across three key directions: adapting pretrained LLMs for communication tasks, developing wireless-specific foundation models to balance versatility and efficiency, and enabling agentic LLMs with autonomous reasoning and coordination capabilities. We highlight recent advances, practical case studies, and the unique benefits of LLM-based approaches over traditional methods. Finally, we outline open challenges and research opportunities—including multimodal fusion, collaboration with lightweight models, and self-improving capabilities—charting a path toward intelligent, adaptive, and autonomous wireless networks.
\end{abstract}

\section{Introduction}

The rapid advancement of large language models (LLMs) has transformed natural language processing, unlocking capabilities in reasoning, representation learning, and generalization from limited supervision. These models, built on transformer architectures and trained on large-scale text corpora, exhibit remarkable adaptability across tasks and domains. As such, their core strengths—sequence modeling, contextual understanding, and zero-shot inference—are increasingly being explored for applications far beyond language, to include robotics, software engineering, and, more recently, wireless communications. This article investigates how LLMs can be strategically repurposed to address key challenges in modern wireless networks, tracing a trajectory from task-specific model adaptation to the realization of autonomous, agent-driven communication systems.

Next-generation wireless systems are characterized by growing complexity and variability. Technologies, such as massive multiple-input multiple-output (MIMO), ultra-dense deployments, and semantic-aware communication, introduce high-dimensional, time-varying environments with stringent latency and reliability constraints. Firstly, these specialized models exhibit poor generalization, often failing to adapt when deployed in new wireless environments with different channel statistics. Secondly, their limited capacity imposes a performance ceiling on increasingly complex tasks where capturing intricate data patterns is essential. Thirdly, their inherent inflexibility makes them prone to overfitting on a single objective, rendering them unsuitable for dynamic, multi-faceted problems. LLMs, by contrast, offer general-purpose reasoning, strong transferability, and cross-task knowledge sharing—making them a promising foundation for more flexible and resilient wireless intelligence. However, harnessing LLMs for wireless tasks is non-trivial. There exists fundamental modality gap between the discrete, semantic token space of LLMs and the continuous, high-dimensional numerical data of wireless systems, such as complex-valued channel state information (CSI) or beamforming vectors. Furthermore, LLMs are typically autoregressive and computationally intensive, posing challenges for low-latency inference in real-time systems. 

To address these issues, recent research has progressed along three key directions. The first focuses on adapting pretrained LLMs to specific wireless tasks, such as beam prediction, channel estimation, and resource allocation. These adaptations involve modality translation, prompt design, and parameter-efficient finetuning strategies, and are shown to outperform traditional models under challenging and mismatched conditions. The second direction aims to develop wireless-specific foundation models—lightweight, multi-task-capable models pretrained on large-scale domain data. These models provide better generalization, reduce training cost, and enable faster adaptation. The third and most forward-looking direction involves the development of agentic LLMs, where LLMs act as autonomous decision-making agents capable of perceiving system state, reasoning over long-horizon goals, coordinating with peers, and invoking external tools, such as simulators.

This article provides a structured overview of these emerging directions. We begin by examining how LLMs can be adapted to support core functionalities in wireless systems, including signal processing, radio resource management, and semantic transmission. We then introduce the concept of wireless foundation models and highlight their advantages in multi-task generalization and domain transfer. Building on these, we explore how LLMs can enable autonomous and distributed coordination through agentic architectures, including multi-agent planning and communication.  We conclude by identifying key challenges and future research directions, such as efficient deployment, multimodal fusion, and continual learning, needed to realize intelligent communication systems.

\section{Adapting LLMs to Wireless Tasks} \label{sec:AdaptLLM}

Although originally developed for natural language processing, LLMs offer powerful sequence modeling and representation learning capabilities that can be repurposed to address core challenges in wireless communication. This section discusses how LLMs have been adapted for key wireless tasks, including physical layer prediction, resource allocation, and semantic communication, by aligning model architecture and data modalities. We also highlight a case study demonstrating the robustness and generalization benefits of LLMs in practice.

\subsection{Physical Layer Prediction Tasks}

Prediction is at the heart of many physical layer operations in wireless systems. Tasks, such as beam selection and channel prediction, rely heavily on anticipating future signal behavior based on past observations. These tasks are inherently sequential and context-dependent—properties that closely align with the pretraining objective of LLMs, which is next-token prediction in a sequence. This fundamental parallel has motivated recent research to explore how LLMs, despite their origins in language processing, can be adapted to perform prediction tasks at the wireless physical layer.

\begin{figure}[htbp]
    \centering
    \includegraphics[width=0.45\textwidth]{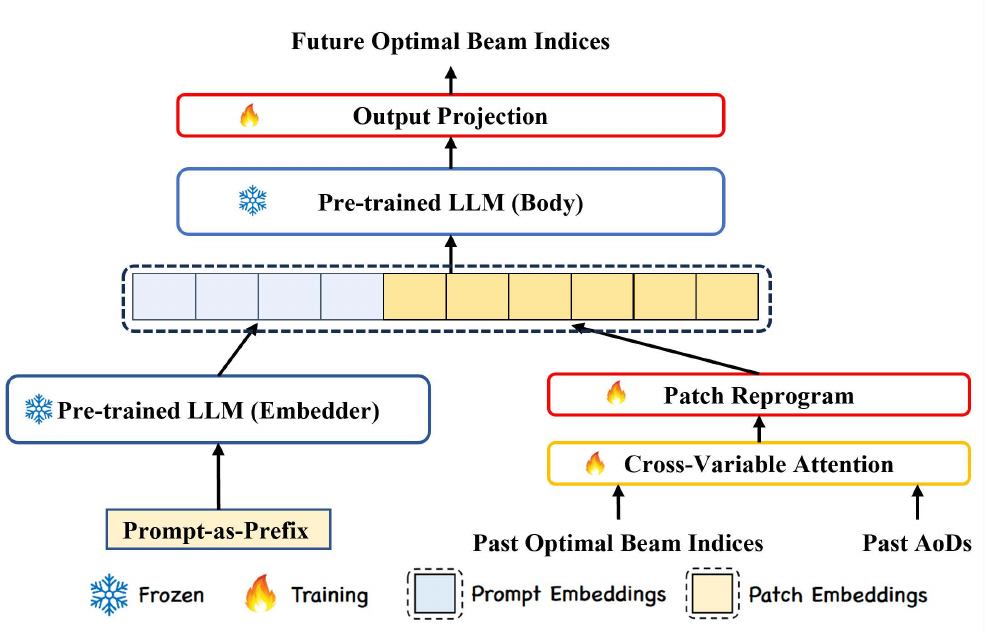}
    \caption{Network architecture of LLM-based beam prediction.}
    \label{fig:beam_structure}
\end{figure}
The key challenge in this adaptation lies in the modality mismatch: standard LLMs are designed to operate on discrete token sequences derived from natural language, whereas wireless data spans a range of formats—from continuous, high-dimensional tensors, e.g., CSI, to discrete but domain-specific labels e.g., beam indices. 
To bridge this gap, recent work has focused on modifying the input and output interfaces of LLMs, leveraging their high-dimensional hidden states to create rich token embeddings that can represent complex signal features with minimal loss of fidelity. As illustrated in Fig.~\ref{fig:beam_structure}, the beam prediction framework  based on LLMs (BP-LLM) \cite{Sheng2025beam} introduces a dual-input strategy: a natural language prompt is used to activate the model’s reasoning capability, while historical beam and angle-of-departure (AoD) sequences are converted into token-like embeddings via a trainable tokenizer. The output is produced through a task-specific linear head designed to predict beam indices. In contrast, to adopt a different approach for channel prediction, the method in \cite{Liu2024channel} bypasses the language interface altogether by injecting learned CSI embeddings directly into the attention layers of the LLM, effectively treating signal features as first-class model inputs. To preserve the generalization benefits of pretraining while minimizing finetuning costs, these adaptations often rely on parameter-efficient tuning strategies, such as low-rank adaptation or adapters. 

Experimental results from the above studies show that adapted LLMs not only achieve competitive performance in full-data regimes but also excel in few-shot and zero-shot settings. Their robustness across varying environments, such as changes in frequency bands or deployment topologies, demonstrates the potential of LLMs to generalize beyond traditional models, which often suffer from brittle performance under distribution shifts. These findings underscore the promise of LLM-based architectures as flexible and generalizable predictors for core physical layer tasks in wireless systems.

Building on these capabilities, recent efforts have explored the idea of consolidating multiple physical layer tasks into a single LLM-based architecture. For example, in \cite{zheng2024llm}, a unified framework has been proposed in which a single LLM backbone, guided by prompt-as-prefix conditioning, is used to perform diverse tasks, such as channel estimation, multi-user precoding, and signal detection. Each task is handled through specialized input and output heads, while the underlying model remains unchanged. This approach illustrates the potential for LLMs to serve not just as task-specific predictors but as a general-purpose physical layer engine, capable of flexibly adapting to new requirements with minimal structural changes.

While the adaptation of LLMs to the physical layer requires overcoming substantial modality mismatches, recent advances in tokenization, embedding design, and efficient finetuning have demonstrated their potential to serve as robust, generalizable predictors for a wide range of physical layer tasks. The primary hurdle to their deployment in real-time systems, inference latency, can be mitigated by promising trends in lightweight architectures, dedicated hardware acceleration, and domain-specific foundation models. This progress, combined with their ability to transfer across environments and tasks, makes them an  alternative to traditional learning-based models in future 6G networks.

\subsection{Resource Allocation}

Resource allocation is central to wireless network optimization, encompassing tasks, such as power control, beamforming, user scheduling, spectrum allocation, etc. These tasks involve selecting actions based on observed network states to achieve predefined objectives, such as maximizing throughput, minimizing energy consumption, or balancing fairness. Traditional approaches to resource allocation typically rely on optimization theory or task-specific deep learning (DL) models, each crafted for a particular combination of input conditions, output actions, and objectives \cite{Pivoto2025resourcesurvey}. Although effective, these solutions are often rigid and lack the flexibility to generalize across varying scenarios.
The emergence of LLMs offers a paradigm shift towards a unified and versatile framework for resource allocation by strong generalization capacity, contextual reasoning, and adaptability. Once provided with an objective, LLMs can be adapted to select specific actions based on network observations, functioning as a decision-making engine.

\begin{figure*}[tbp]
    \centering
    \includegraphics[width=1.0\textwidth]{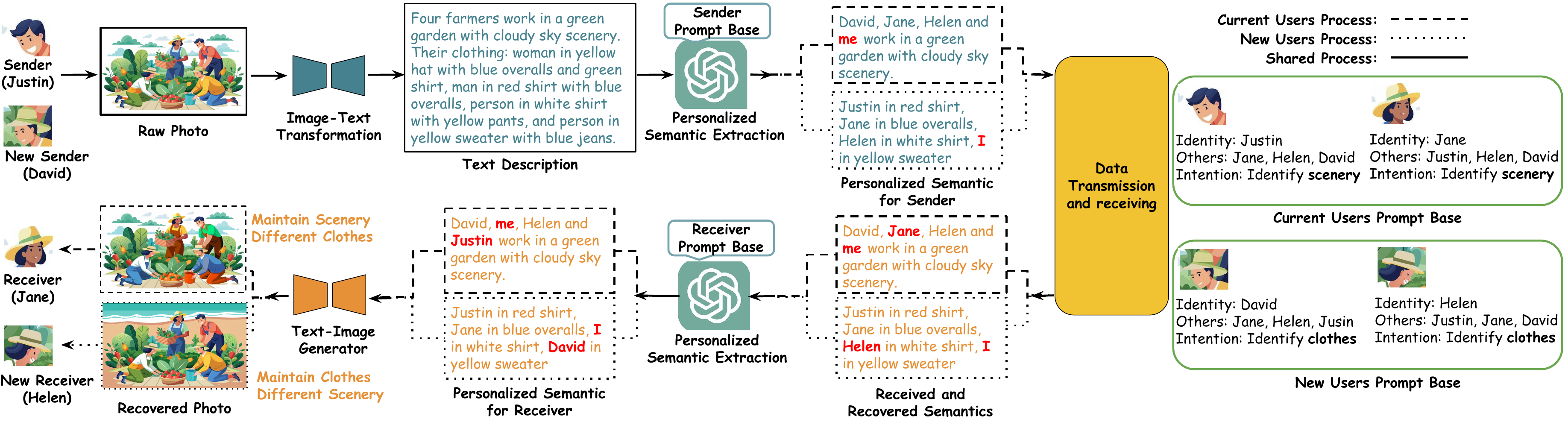}
    \caption{Illustration of personalized semantic extraction via LLM in the semantic communication system for group photo transmission. Currently, Justin (sender) transmits this photo to Jane (receiver), and both of them intend to identify people and background in this photo. In the sender side, the LLM uses their prompt bases to extract intention-aligned semantics (omitting irrelevant details like clothing colors). Furthermore, the LLM transforms sender-centric references to receiver-perspective semantics (``me'' to ``Justin'', ``Jane'' to ``me''). Due to their intention only on people and background, clothing colors in the recovered image differ from the original. New users David and Helen can seamlessly join in this system by providing their prompts focused on clothing. The LLM now prioritizes clothes details over background, yielding a recovered image for Helen that preserves original clothing colors but alters the background. }
    \label{fig:semantic_user}
\end{figure*}

On the input side, network observations, such as CSI matrices, radar point clouds, and camera feeds, must be converted into representations comprehensible to the LLMs. This conversion is achieved using hybrid architectures where modality-specific encoders transform diverse inputs into token-like embeddings \cite{yin2024a}. These embeddings are subsequently fed into the LLM backbone. The LLMs then leverage the pretrained capacity for contextual reasoning and feature extraction to generate rich, generalizable feature representations.

Following input processing, the output heads map the feature representations to specific resource allocation actions, such as transmit power, precoding vectors, and scheduling indicators. However, the sequential output process of LLMs introduces two critical issues for real-time applications: significant inference latency and the risk of generating invalid outputs that fall outside the defined action space (``hallucinations''). To address this, architectural innovations, such as NetLLM, replace the language-based output head with a task-specific classifier, which maps internal representations directly to the vaild action space \cite{wu2024netllm}. This design enables the model to output a valid action in a single inference step, thereby meeting the strict latency and reliability demands of real-time systems.

Regarding objective understanding, a key advantage of LLMs is their ability to interpret natural language directives, which enables dynamic goal alignment. These models can be guided using techniques such as prompt-as-prefix, where a target objective is provided as a natural language component of the input. Through task-specific training methods, like reinforcement learning (RL), the LLMs then adapt their policy to align with the provided objective. While effective only for objectives encountered during training, it lays the groundwork for future agentic systems capable of interpreting unseen instructions, planning, and adapting autonomously. This topic will be further explored  in Section \ref{sec:agentic LLM}.

\subsection{Semantic Communication}

Semantic communication seeks to transmit the meaning of information rather than its raw representation, aiming to reduce redundancy and improve communication efficiency. Conventional DL-based semantic communication systems adopt an autoencoder architecture, where the encoder compresses input into semantic features and the decoder reconstructs the original content from distorted features received over the channel. However, these systems suffer from poor robustness to channel variations and lack flexibility. They are typically trained for specific tasks or users, and adapting to new scenarios requires retraining the entire encoder-decoder pipeline—posing scalability challenges for dynamic, heterogeneous networks.

LLMs offer a promising alternative by serving as a shared knowledge base that supports task reasoning, interpretable semantic representation, and generalization across tasks and users. When given a task prompt, LLMs employ chain-of-thought reasoning to decompose complex instructions into subtasks, select standardized semantic formats, and guide the construction of semantic encoders and decoders accordingly. This modular design decouples semantic processing from channel adaptation, allowing intermediate components to handle distortions without retraining the encoder or decoder.

This paradigm shift enables greater flexibility. For example, in the 
multi-user generative semantic communication framework \cite{yang2025rethinking}, the LLM can interpret the task as requiring an object-focused scene layout when prompted with a scene reconstruction task. In the large AI model-based multimodal semantic communication system \cite{jiang2024large}, user-specific prompts guide semantic extraction and recovery, supporting personalized semantics based on identity, interests, and intent. As shown in Fig.~\ref{fig:semantic_user}, the LLM adjusts both the encoded content and receiver interpretation using personalized prompt bases, allowing multiple users to perceive the same message differently based on their preference.

These systems also highlight a key strength of LLMs: the ability to extract different semantic features for different tasks from the same input, without retraining. For instance, given a single image, prompt tuning allows the LLM to switch between object detection and scene captioning, tailoring the extracted semantics accordingly. Traditional DL systems would require separate training pipelines for each. Despite these advantages, LLM-powered semantic communication faces a major bottleneck: computational cost. Running a full LLM on every task instance is infeasible at the network edge. A more practical strategy is to execute the LLM once in the cloud per task, generating meta-instructions or latent representations to guide lightweight local semantic communication modules. This hybrid setup balances the generalization power of LLMs with the real-time efficiency required by edge networks.

\begin{figure}[htbp]
    \centering
    \includegraphics[width=0.4\textwidth]{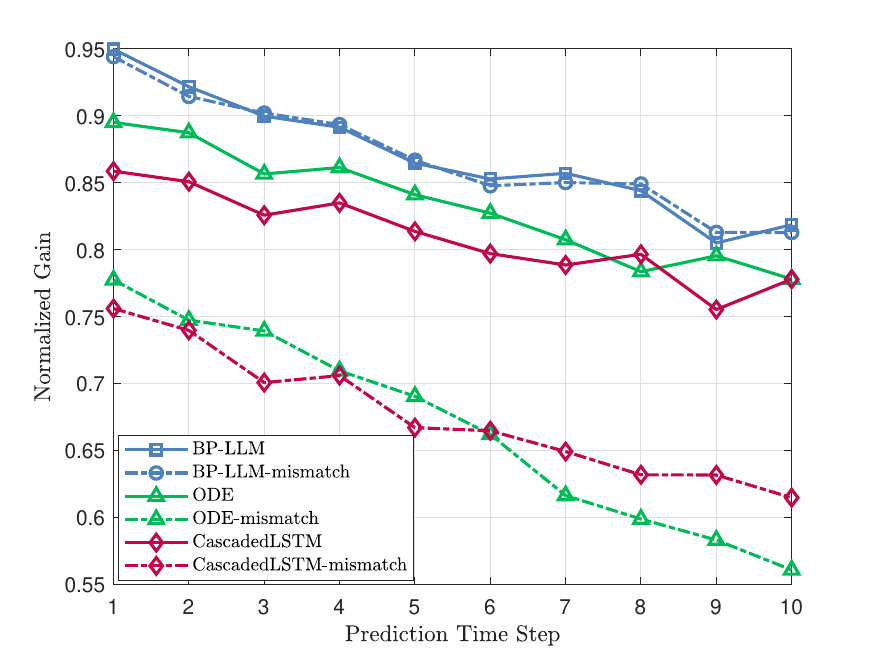}
    \caption{Beam prediction performance of the BP-LLM method \cite{Sheng2025beam} compared with other learning-based baselines under the mismatched scenarios.}
    \label{fig:beam_mismatch}
\end{figure}

\subsection{Case Study: BP-LLM for Robust Beam Prediction}
To concretely demonstrate the adaptation of LLMs to physical layer tasks,  we highlight BP-LLM \cite{Sheng2025beam}, a model designed to predict optimal beam directions based on historical beam and AoD sequences. As shown in Fig.~\ref{fig:beam_mismatch}, BP-LLM is tested under mismatched conditions (e.g., trained at 28 GHz, tested at 60 GHz) and compared to LSTM baselines. The LSTM models experience sharp performance degradation but BP-LLM remains robust—validating the generalization power of LLMs in real-world deployments. This case study underscores the promise of LLM-based architectures for robust and adaptive wireless communications.

\section{Wireless Foundation Model} \label{sec:WFM}

While adapting LLMs to communication tasks can yield strong performance and enhance generalization across tasks, several limitations remain. In particular, the large parameter size and high inference latency of general-purpose LLMs hinder their applicability in delay-sensitive wireless systems, especially at the physical layer. This motivates the development of wireless-specific foundation models. These compact, domain-specific models are pretrained on large-scale communication data and retain the advantages of LLM adaptation while improving efficiency, robustness, and scalability.

The concept of foundation models has reshaped DL in recent years \cite{bommasani2021fOundation}, emphasizing general-purpose architectures pretrained on diverse data from a target domain. Unlike LLMs, wireless foundation models are not defined by size alone. Rather, their key advantage lies in their ability to encode domain-specific knowledge through pretraining, enabling few-shot, zero-shot, and multi-task capabilities across different communication scenarios, while maintaining fast inference. In what follows, we examine the development of wireless foundation models for two core areas: physical layer prediction and resource management.

\subsection{Physical Layer Foundation Models}

The predictive nature of physical layer tasks, such as channel estimation and beam selection, makes them well-suited for foundation models trained via self-supervised learning. While Section \ref{sec:AdaptLLM} introduced LLM-based approaches that adapt autoregressive architectures for these tasks, a parallel line of work explores leaner and faster alternatives.

A representative example, WiFo \cite{liu2024wifo}, introduces a masked autoencoder (MAE) framework to build a self-supervised foundation model for channel prediction. The model is pretrained on CSI data across spatial, temporal, and frequency domains using a masking strategy that forces it to infer missing entries. Compared to LLM-based solutions \cite{Liu2024channel}, WiFo achieves lower inference latency but superior accuracy across a variety of deployment scenarios, demonstrating the efficiency advantage of task-specific architectures.

Beyond channel prediction, foundation models have been shown to support multiple physical layer tasks in a unified framework. WirelessGPT \cite{yang2025wirelessgpt} leverages unsupervised learning to extract shared CSI embeddings, which are then reused across tasks, such as beamforming, signal detection, and modulation classification. This eliminates the need for separate models for each function and enables smoother adaptation to new physical layer requirements with task-specific heads. In contrast to tightly coupled, task-specific systems, foundation models offer a modular and reusable design that aligns with the trend toward composable and efficient wireless architectures.

\subsection{Predictive Foundation Models for Resource Management}

While Section \ref{sec:AdaptLLM} discussed LLMs for resource allocation as decision-makers, foundation models serve a different but equally critical role: generating predictive inputs that drive better decisions. Resource management in wireless systems—including user scheduling, handovers, and power control—depends heavily on accurate forecasting of channel conditions, user mobility, and traffic demand \cite{Pivoto2025resourcesurvey}. However, existing predictive models are often scenario-specific, exhibiting limited generalization to new environments and struggling with long-horizon predictions. 

To close this gap, a unified foundation model has been proposed in \cite{sheng2025wfm} for multi-task prediction in wireless networks that supports diverse intervals. The predictive foundation model enforces univariate decomposition to unify heterogeneous tasks, encodes granularity for interval awareness, and uses a causal transformer backbone for accurate predictions. Additionally, a patch masking strategy is introduced during training to support arbitrary input lengths. Trained on large datasets, the predictive foundation model can jointly predict future CSI, angle-of-arrival (AoA), and network traffic, supporting multi-step forecasting for proactive resource allocation. Their generalization capabilities also extend to novel tasks without retraining, offering strong zero-shot performance.

More importantly, these predictions can be seamlessly integrated into RL frameworks. In model-free RL, the predicted multi-step future information can be embedded into the agent’s state representation, enabling optimization for long-term objectives like energy efficiency over short-term utility. In model-based RL, these predictive trajectories allow for simulating future states to evaluate different actions. For example, Monte Carlo tree search (MCTS) \cite{coulom2006MCTS} can leverage these predictions to plan multiple steps ahead, which is particularly useful in wireless systems where control overhead and latency are tightly constrained. This synergy between predictive foundation models and RL enables the development of intelligent and scalable solutions for adaptive resource management in complex, dynamic environments.

\subsection{Case Study: Zero-Shot Generalization in Multi-Task Predictive Foundation Model}

\begin{figure}[htbp]
    \centering
    \includegraphics[width=0.4\textwidth]{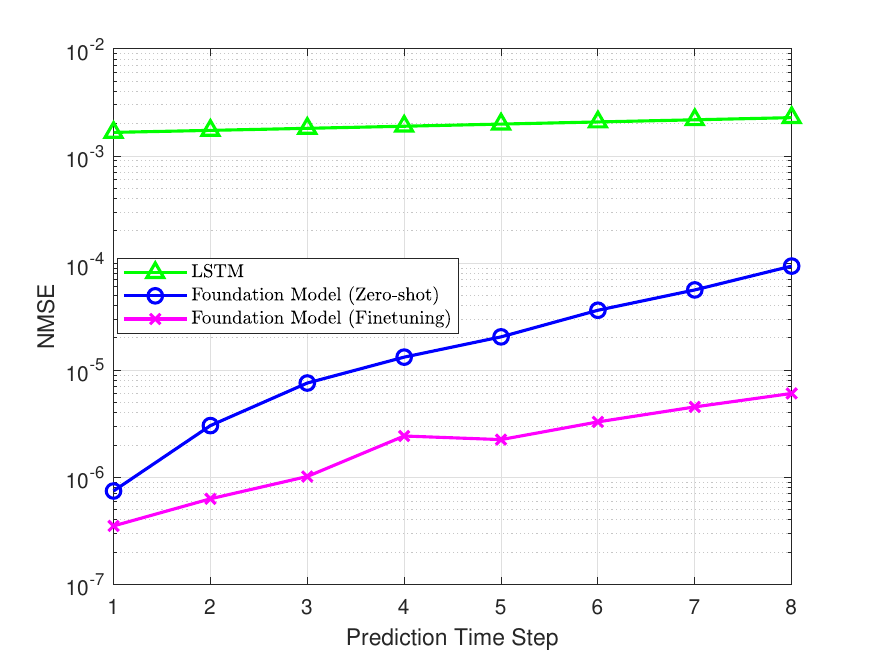}
    \caption{The zero-shot prediction performance of the wireless foundation model compared with the full-shot baselines under the user distance prediction task.}
    \label{fig:wpfm_unseen_task}
\end{figure}

This case study investigates the wireless foundation model in \cite{sheng2025wfm} that performs multi-task prediction for three different variables of interest: CSI, AoA, and network traffic. To evaluate its zero-shot generalization capability, we directly test the model on an unseen task: predicting user distance, without any additional training.  As shown in Fig.~\ref{fig:wpfm_unseen_task}, the foundation model (“Foundation Model (Zero-shot)”) outperforms the standard full-shot LSTM and Transformer baselines, trained under the user distance prediction task, in normalized mean-squared error (NMSE), despite never being trained on the distance prediction task. This superior zero-shot performance stems from the model's ability to capitalize on the inherent structural similarities among these time-series prediction tasks. Extensive pre-training yields a generalized understanding of temporal dynamics, which proves directly applicable to the unseen distance prediction problem. After minimal finetuning (“Foundation Model (Finetuning)”), performance improves further. These results highlight the strong zero-shot learning capabilities of the wireless foundation model.

\section{Agentic LLM for Wireless Communication}\label{sec:agentic LLM}

The preceding sections have shown how pretrained LLMs and wireless foundation models can enhance communication functions, such as semantic compression, channel prediction, and resource management. These advances primarily address well-defined tasks with clear inputs, outputs, and performance goals. However, to manage the immense complexity of future 6G networks, a paradigm shift is required from creating task-specific tools to engineering autonomous systems. This marks the final stage in the evolution from adaptation to autonomy: endowing LLMs with agency—the ability to perceive, reason, and act autonomously in dynamic environments. While the foundational concepts of agentic AI are rooted in general computer science, our focus here is their integration into wireless systems to achieve holistic network self-organization.

\subsection{Key Capabilities of Agentic LLMs}

Unlike traditional approaches such as rule-based systems or pure RL agents, which often lack flexibility or require extensive environment-specific training, agentic LLMs overcome these limitations by combining the reasoning ability of symbolic systems, the adaptability of RL, and the vast prior knowledge embedded within LLMs.

Crucially, their decision-making policies are often not learned from pre-existing, static datasets. Instead, agentic LLMs can operate through two primary modes: i) in-context learning, where strategies are guided by detailed prompts, and ii) online interaction with the environment (or a simulator). In the latter mode, agents generate their own streams of experience and adapt their policies via reflection, a paradigm recently explored for intelligent network management \cite{tong2025wirelessagent}. The effectiveness of this approach in wireless communication is underpinned by three core capabilities:

\begin{itemize}

\item \textbf{Reasoning and Planning:} 
The ability to parse high-level goals, decompose them into a sequence of executable subgoals, and formulate a multi-step action plan. This core capability enables them to devise strategies in zero-shot settings by leveraging their pretrained knowledge.

\item \textbf{Memory and Reflection:} 
The capacity to learn from past interactions by utilizing a dual memory system: short-term memory for immediate context and long-term memory for accumulated knowledge \cite{lewis2020rag}. This allows agents to reflect on past actions, identify errors, and continually refine their strategies over time.

\item \textbf{Tool Use:} The functionality to interact with external environments through standardized interfaces. This enables the invocation of specialized tools—such as channel simulators, resource allocation algorithms, and other APIs—thereby extending the agent’s perceptual and actuation capabilities beyond its internal knowledge.

\end{itemize}

Collectively, these key capabilities position agentic LLM as a promising paradigm for building intelligent, flexible, adaptive and autonomous wireless networks.

\subsection{Applications in Wireless Systems}

Recent work has demonstrated the potential of agentic LLMs in practical wireless applications. The WirelessAgent prototype \cite{tong2025wirelessagent} showcases how an agentic LLM can infer user intent, optimize in-slice beamforming, and dynamically manage network slicing to meet quality of service requirements. This illustrates a pathway toward self-organizing networks, where agentic LLMs operate as a centralized controller with comprehensive network awareness, memory of traffic patterns, and predictive capabilities for future service demands.

This centralized model unifies traditionally fragmented, cross-layer tasks into a single decision-making framework. The agent interprets natural language prompts specifying cross-layer constraints, including latency budgets, energy limits, and service priorities. It then leverages tools, such as channel simulators, to evaluate potential configurations. Through reflection, the agent iteratively refines its strategy, converging on high-quality solutions without the extensive environment interaction demanded by conventional RL approaches. However, as wireless networks scale, this centralized architecture encounters critical bottlenecks in latency, computation, and fault tolerance. These limitations necessitate collaborative multi-agent designs, which typically follow two structural paradigms.

In vertical architectures, a leader agent aggregates local reports, performs global reasoning, and dispatches control commands. This top-down model supports fast convergence and clear coordination but introduces a single point of failure and limited scalability tied to the lead agent’s capacity.  In contrast, a horizontal architecture distributes autonomy by allowing all agents to share a communication interface, broadcast observations, negotiate sub-tasks, and invoke tools independently. This peer-to-peer structure enhances fault tolerance and scalability, making it suitable for decentralized tasks like federated learning and interference coordination. However, it can suffer from excessive communication overhead unless optimized with strategies such as message compression or clustering. Integrating these architectural patterns is crucial for developing resilient and scalable multi-agent systems to power next-generation intelligent wireless networks.

\subsection{Case Study: Multi-AP Coordination with Agentic LLMs}

\begin{figure}[htbp]
    \centering
    \includegraphics[width=0.49\textwidth]{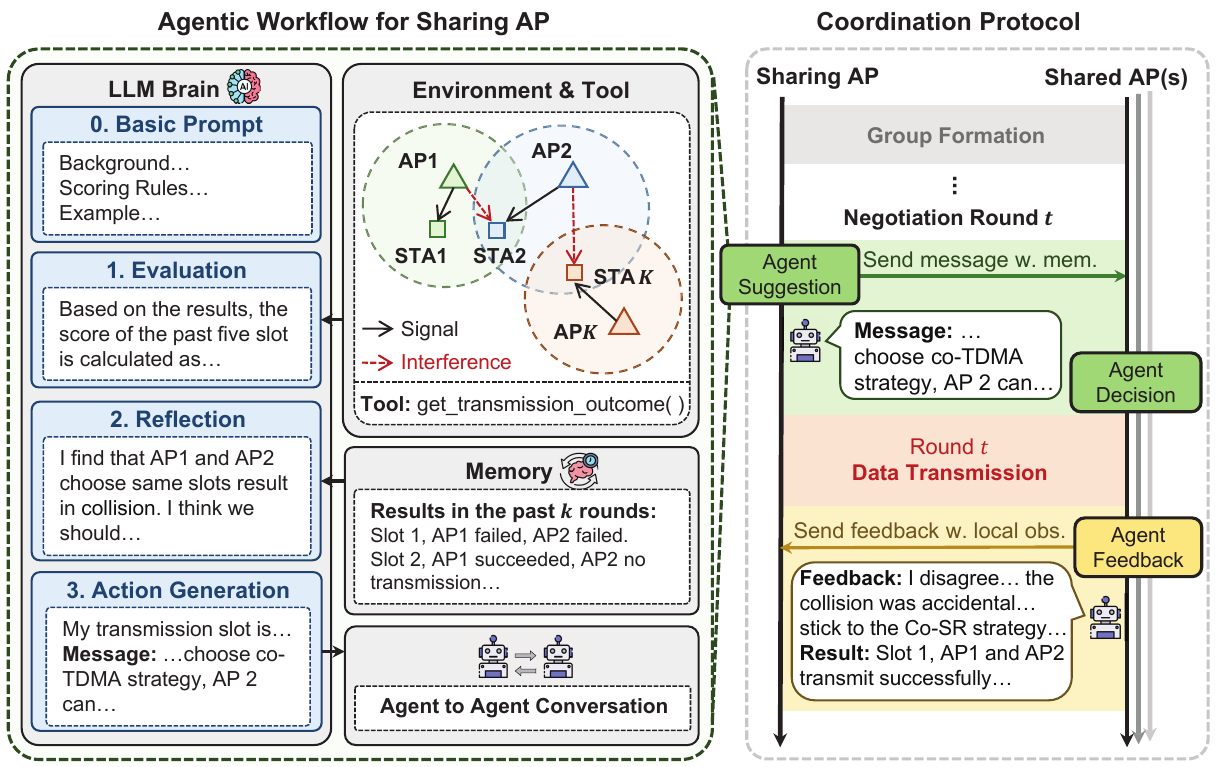}
    \caption{Agentic workflow and coordination protocol for multi-AP cooperation.}
    \label{fig:m_ap_coordination}
\end{figure}

This case study illustrates a decentralized vertical multi-agent framework using agentic LLMs for coordinated transmission in the Wi-Fi overlapping basic service set (OBSS) environments. Each access point (AP) is modeled as an autonomous LLM agent capable of natural language communication, reasoning, and decision-making, thereby enabling adaptive cooperation in dense wireless deployments.

Recent updates in Wi-Fi 8 introduce standardized support for multi-AP coordination, including mechanisms, such as coordinated time division multiple access (Co-TDMA) and coordinated spatial reuse (Co-SR). These mechanisms allow APs to share critical information to mitigate interference between basic service sets (BSSs). In our agentic framework, these coordination procedures are facilitated through communication among AP agents, allowing them to negotiate and learn adaptive channel access strategies.

Figure~\ref{fig:m_ap_coordination} outlines the agentic workflow for sharing AP and the coordination protocol for multi-AP cooperation. Internally, each agent continuously refines its strategy through a cognitive loop of evaluation and reflection. Leveraging memory of past outcomes to evaluate performance, the agent reflects on the outcomes of its actions (e.g., transmission collisions) to proactively refine its strategy. External tools ground this internal process in the physical world. Externally, this reflective capability enables a sophisticated coordination protocol where agents engage in multi-round negotiation and feedback, allowing them to collectively adapt to the environment rather than adhering to rigid, pre-defined rules.

\begin{figure}[htbp]
    \centering
    \includegraphics[width=0.43\textwidth]{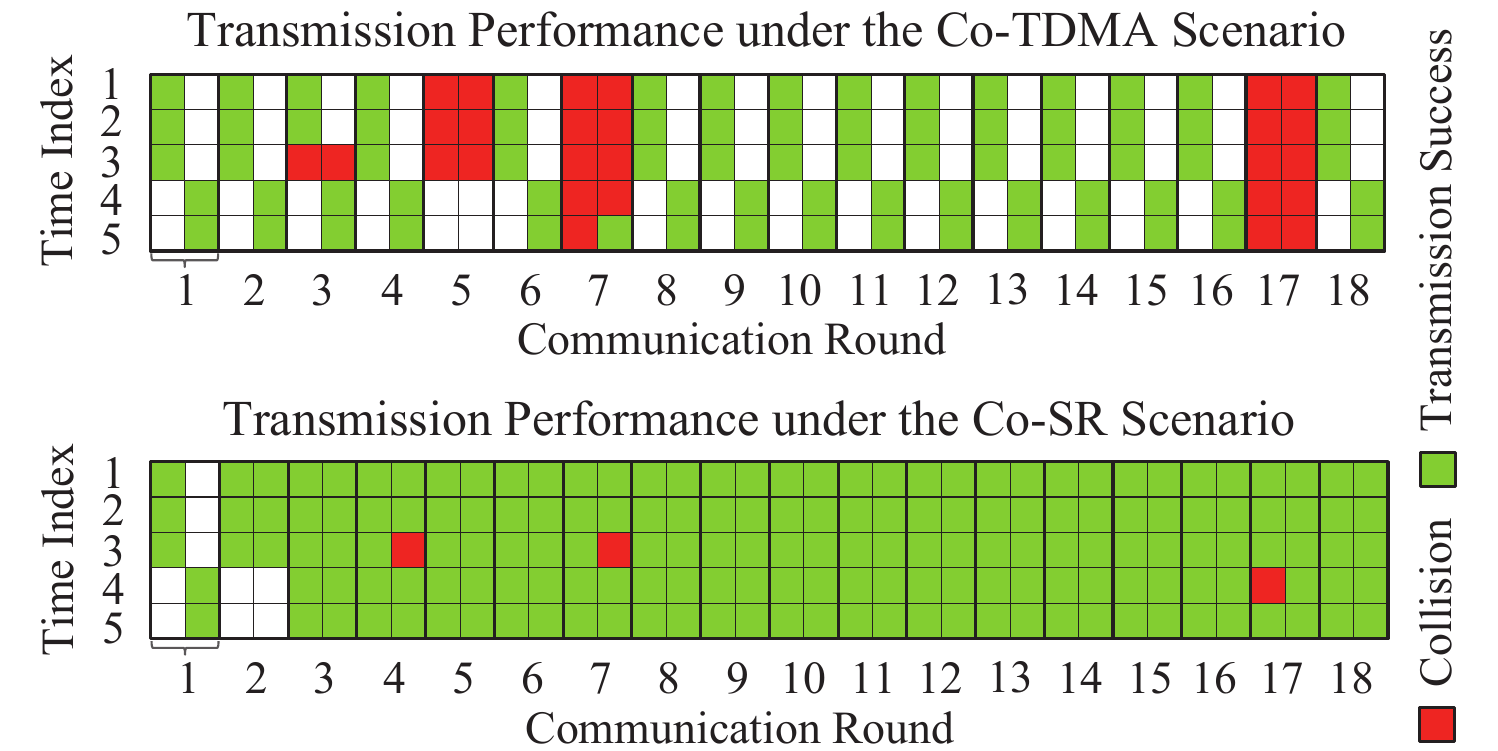}
    \caption{Transmission performance of the agentic multi-AP coordination protocol under the Co-TDMA and Co-SR scenarios. For each communication round, the two columns within the bold frame display the transmission states of the sharing AP (left) and the shared AP (right), respectively.}
    \label{fig:agentic_result}
\end{figure}

Simulation results in Fig. \ref{fig:agentic_result} validate the effectiveness of the proposed protocol. In the Co-TDMA scenario, the agents initially experience several collisions. However, by leveraging the reflective and negotiation processes detailed in Fig. \ref{fig:m_ap_coordination}, they rapidly converge on a stable, collision-free transmission schedule after a few rounds. In the Co-SR scenario, the agents progressively explore more aggressive strategies to maximize throughput. This ability to autonomously discover and adapt coordination strategies, without reliance on pre-programmed heuristics, underscores the potential of agentic LLMs to manage dynamic and complex wireless environments.

\section{Challenges and Opportunities}

Despite their significant promise, the application of LLMs in communications faces considerable challenges. This section outlines key directions for future research.

\subsection{Collaboration of LLM and Small Models}
 LLMs exhibit powerful generalization and reasoning capabilities but incur high computational cost and latency—making them ill-suited for real-time or resource-constrained tasks, particularly in the physical layer processing. A promising direction lies in designing collaborative frameworks where LLMs act as high-level coordinators while small domain-specific models handle fast, localized tasks, such as beam prediction, channel estimation, and power control. This hierarchical architecture can balance global reasoning with real-time responsiveness. Future research should also explore dynamic task delegation mechanisms, where LLMs selectively invoke small models based on context and performance goals.

\subsection{Multimodal Integration for Wireless Foundation Models}
Current wireless foundation models are predominantly unimodal, limiting their ability to handle heterogeneous data sources, such as images, text, and sensory data (radar, LiDAR, etc.), that coexist in intelligent environments. Enabling multimodal processing in wireless foundation models can significantly enhance situational awareness and decision-making. A unified representation learning framework, mapping different modalities into a shared semantic space, can support cross-modal reasoning and reduce modality mismatch. Addressing challenges, such as modality alignment, temporal synchronization, and robustness to missing modalities, is essential for deploying these systems in real-world wireless settings.

\subsection{Safety and Verification of Agentic LLMs}
The generative and autonomous nature of agentic LLMs poses significant challenges in guaranteeing safe and predictable behavior in mission-critical networks. Existing reactive approaches, such as safety guardrails, can prevent basic violations on a per-action basis but are insufficient for ensuring the logical integrity of complex, multi-step plans. A promising direction lies in adopting formal verification, a proactive paradigm that can mathematically prove a plan's adherence to critical system invariants before execution. Future research should develop scalable verification algorithms for long-horizon tasks and robust methods for translating ambiguous, LLM-generated plans into verifiable formal models.

\subsection{Non-Linguistic Agentic Large AI Models}
Agentic LLMs currently generate decisions through natural language output, which introduces fairly high inference delays. In real-time wireless systems, this latency can be a critical bottleneck. A promising direction is to develop non-linguistic agentic models that preserve the high-level reasoning and planning capabilities but express decisions through more compact representations, such as structured commands, vector embeddings, or symbolic actions. By bypassing verbose natural language output, these models sharply reduce decision latency, making it more suitable for time-sensitive control.

\section{Conclusion}
The integration of LLMs into wireless communications marks a transformative step toward building more intelligent, adaptive, and efficient networks. From adapting pretrained LLMs for physical layer prediction and resource allocation, to developing compact wireless foundation models, and finally to enabling agentic LLMs capable of autonomous reasoning and coordination—each stage brings us closer to self-optimizing communication systems. These models offer strong generalization, multi-tasking, and zero-shot capabilities that address many of the core limitations in traditional and DL-based wireless solutions. However, realizing their full potential will require tackling key challenges, such as latency, model efficiency, multimodal integration, and lifelong learning. As 6G and beyond continue to redefine the demands placed on wireless infrastructure, LLM-driven systems offer a promising path toward networks that are not only more capable—but also more intelligent, flexible, and autonomous.

\bibliographystyle{IEEEtran}

\bibliography{reference.bib}

\vspace{-35pt}
\begin{IEEEbiographynophoto}{Le Liang}
is a Professor with the School of Information Science and Engineering, Southeast University, China. His main research interests are in wireless communications and machine learning. 
\end{IEEEbiographynophoto}

\vspace{-35pt}
\begin{IEEEbiographynophoto}{Hao Ye}
is an Assistant Professor in the Department of Electrical and Computer Engineering at the University of California, Santa Cruz. His main research interests include machine learning and wireless communications.
\end{IEEEbiographynophoto}

\vspace{-35pt}
\begin{IEEEbiographynophoto}{Yucheng Sheng}
is currently pursuing the Ph.D. degree in the School of Information Science and Engineering at Southeast University, China. His research interests include LLMs and foundation models for wireless communication.
\end{IEEEbiographynophoto}

\vspace{-35pt}
\begin{IEEEbiographynophoto}{Ouya Wang}
is currently pursuing the Ph.D. degree with the Department of Electrical and Electronic Engineering at Imperial College London, UK. His research interests include accretionary learning and deep learning with application in signal processing.
\end{IEEEbiographynophoto}

\vspace{-35pt}
\begin{IEEEbiographynophoto}{Jiacheng Wang}
is currently pursuing the Ph.D. degree in the School of Information Science and Engineering at Southeast University, China. His research interests include learning-based resource allocation for wireless communication.
\end{IEEEbiographynophoto}

\vspace{-35pt}
\begin{IEEEbiographynophoto}{Shi Jin}
is Vice President and a Professor with Southeast University, China. His main research interests include wireless communications, random matrix theory, and information theory. He is an IEEE and IET Fellow. 
\end{IEEEbiographynophoto}

\vspace{-35pt}
\begin{IEEEbiographynophoto}{Geoffrey Ye Li}
is a Chair Professor at Imperial College London, UK. He was elected to Fellow of the Royal Academy of Engineering (FREng), IEEE Fellow, and IET Fellow for his contributions to signal processing for wireless communications.
\end{IEEEbiographynophoto}

\end{document}